%% file: ElementsOfML.tex
\documentclass[journal]{IEEEtran}
\usepackage[utf8]{inputenc} 
\usepackage[T1]{fontenc}    
\usepackage{hyperref}       
\usepackage{url}            
\usepackage{booktabs}       
\usepackage{amsfonts}       
\usepackage{nicefrac}       
\usepackage{microtype}      
\usepackage{csvsimple}
\usepackage{amsopn}
\usepackage{amsthm}
\usepackage{etex}
\usepackage{graphicx}
\usepackage{endnotes}
\usepackage{hyperref}
\usepackage{epsfig,psfrag}
\usepackage{pst-all}
\usepackage{amssymb,amsfonts,upref,cite,epsf,color,bm}
\usepackage{graphicx}
\usepackage{color}
\usepackage{amsmath}
\usepackage{graphicx}
\usepackage{calc}
\usepackage{booktabs}
\usepackage{tikz}
\usepackage{pgfplots}
\usepackage{tikz}

\input{mlbpbook_macros}

\definecolor{aqua}{rgb}{0.0, 1.0, 1.0}

\usepackage{algorithm}
\usepackage{algpseudocode}
\floatname{algorithm}{Algorithm}
\algnewcommand\algorithmicinput{\textbf{Input:}}
\algnewcommand\INPUT{\item[\algorithmicinput]}
\algnewcommand\algorithmicoutput{\textbf{Output:}}
\algnewcommand\OUTPUT{\item[\algorithmicoutput]}

\newcommand{\featurelen}{\featuredim}

\usepackage{setspace}
\usetikzlibrary{arrows,positioning} 
\tikzset{
    >=stealth',
    punkt/.style={
           rectangle,
           rounded corners,
           draw=black, very thick,
           text width=3cm,
           minimum height=2em,
           text centered},
    pil/.style={
           ->,
           thick,
           shorten <=2pt,
           shorten >=2pt,}
}

\usetikzlibrary{arrows,automata,snakes}
\usetikzlibrary{calc,trees,positioning,arrows,chains,shapes.geometric,%
decorations.pathreplacing,decorations.pathmorphing,shapes,%
matrix,shapes.symbols,automata}

\tikzset{
  treenode/.style = {shape=rectangle, rounded corners,
                     draw, align=center,
                     top color=white, bottom color=blue!20},
  root/.style     = {treenode, font=\Large, bottom color=red!30},
  env/.style      = {treenode, font=\ttfamily\normalsize},
  dummy/.style    = {circle,draw}
}

\title{Components of Machine Learning:\\ Binding Bits and FLOPS}

\author{Alexander Jung 
\thanks{Author is with the Department of Computer Science, Aalto University, Finland; firstname.lastname(at)aalto.fi}
}

\begin{document}
	\maketitle
\begin{abstract}
Many machine learning problems and methods are combinations of three components: 
data, hypothesis space and loss function. Different machine learning methods 
are obtained as combinations of different choices for the representation of data, 
hypothesis space and loss function. After reviewing the mathematical structure 
of these three components, we discuss intrinsic trade-offs between statistical 
and computational properties of machine learning methods.  
\end{abstract}

\section{Introduction}
\label{sec_intro}
Machine learning (ML) methods implement the scientific principle of continuous verification and 
adaptation of a hypothesis about an observable phenomenon (``observable fact or event'') \cite{PopperBook}. 
Examples of a phenomena are: 
\begin{itemize}
\item the visual scene recorded by the smartphone snapshot depicted in Figure \ref{fig:image}.
\item the hiking time required to reach the peak in Figure \ref{fig:image}. 
\item the water temperature of the lake in Figure \ref{fig:image}. 
\end{itemize} 
The verification and adaption of the hypothesis is based on the observation of data. ML theory and 
methods revolve around the implementation of the cycle underlying this principle using limited 
computational resources such as computation time and storage capacity. 

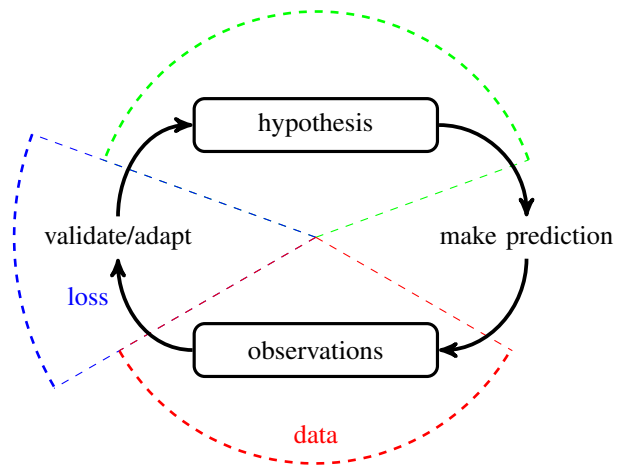
\begin{figure}[htbp]
\hspace*{5mm}
\begin{tikzpicture}[node distance=1cm, auto,]
\coordinate (OR) at (0.00, 1.50);
  \node[circle,inner sep=0,minimum size={6cm}](a) at (OR) {};
    \node[circle,inner sep=0,minimum size={8cm}](b) at (OR) {};
        \node[circle,inner sep=0,minimum size={6cm}](c) at (OR) {};
\draw[red,line width=1,dashed] (a.-150) arc (-150:{-150+120}:3cm);
\draw[blue,line width=1,dashed] (b.160) arc (160:{160+50}:4cm);
\draw[green,line width=1,dashed] (c.20) arc (20:{160}:3cm);
   \draw[green,thin,dashed] (OR) -- (c.20);
      \draw[green,thin,dashed] (OR) -- (c.160);
   \draw[blue,thin,dashed] (OR) -- (b.160);
      \draw[blue,thin,dashed] (OR) -- (b.210);
   \draw[red,thin,dashed] (OR) -- (a.-150);
   \draw[red,thin,dashed] (OR) -- (a.-30);   
 \node[punkt] (data) {observations};
  \node[red,below=0.5cm of data] (data1) {data};
 \node[above=of data] (dummy) {};
  \node[punkt,above=1cm of dummy] (hypothesis) {hypothesis};
 \node[right=1.4cm of dummy] (t) {make prediction} ; 
 \node[left=1.4cm of dummy] (g) {validate/adapt} ; 
  \node[blue,below=0.5cm of g,anchor=east] (g1) {loss} ; 
 \draw [->,line width=0.5mm] (hypothesis.east) to [out=0,in=90] (t.north);
  \draw [->,line width=0.5mm] (t.south) to [out=270,in=0] (data.east);
    \draw [->,line width=0.5mm] (data.west) to [out=180,in=270] (g.south);
        \draw [->,line width=0.5mm] (g.north) to [out=90,in=180] (hypothesis.west);
\end{tikzpicture}
 \vspace*{-3mm}
\caption{The cycle of the scientific principle which is implemented by ML methods. 
Main components of ML methods are data, a hypothesis space and a loss function.}
\label{fig_cycle_scientific_principle}
\end{figure}
Modern ML methods execute the cycle in Figure \ref{fig_cycle_scientific_principle} within a 
fraction of a second and using billions of data points \cite{Goodfellow-et-al-2016}. Deep Learning 
methods implement the cycle of Figure \ref{fig_cycle_scientific_principle} by representing  
hypotheses by artificial neural networks whose weights (parameters) are continuously 
adapted using (variants of) gradient descent \cite{Goodfellow-et-al-2016}. 

A typical ML method consists of three components: 
\begin{itemize}
\item data (mostly in the form of a huge number of bits)
\item a hypothesis space (also referred to as a ML model) consisting of computationally feasible predictor functions. 
\item a loss function that is used to assess the quality of a particular predictor function. 
\end{itemize}

To implement ML methods, given a limited amount of computational resources such as number of floating point operations per second (FLOPS), 
we need to be able to efficiently store and manipulate data and predictor functions.
One extremely efficient approach to represent and manipulate data and predictor functions 
are matrices and vectors. The mathematical foundation of computing with matrices and vectors is linear algebra \cite{StrangLinAlg2016}. Therefore, 
a large part of ML theory and methodology is applied numerical linear algebra.

Indeed, data points can often characterized by a list of numeric attributes $x_{r}$ which can be stacked 
into a vector\footnote{We use bold font to represent vectors such as $\mathbf{x}$ or $\mathbf{w}$.} $\mathbf{x}=\big(x_{1},\ldots,x_{n}\big)^{T}$. Moreover, many ML methods (such as linear regression 
or logistic regression) use predictor functions of the form $h(\mathbf{x}) = \sum_{r=1}^n w_r x_r  = \mathbf{w}^{T} \mathbf{x}$ with some weight vector 
$\mathbf{w}=(w_{1},\ldots,w_{n})^{T}$. Note that once we restrict ourselves to linear functions of the form $h(\mathbf{x}) = \mathbf{w}^{T} \mathbf{x}$, 
we can represent a predictor function by the weight vector $\mathbf{w}$. Indeed, given the weight vector $\mathbf{w}$, 
we can evaluate the predictor function for any feature vector $\mathbf{x}$ as $h(\mathbf{x}) = \mathbf{w}^{T} \mathbf{x}$.  

\section{Data}

The key component of any machine learning problem (and method) is data. There are many 
different sources of data such as text documents, sensor measurements, videos or image collections. 
Digital data is available in the form of a stream of bits which needs to be parsed into elementary 
units which represent individual data points. Data points might be represented by rows in a spreadsheet, 
the set of weather observations in Finland during a specific period of time, images, audio recordings or 
entire digital footprints of humans. 

Typically, we have never full access to (every single detailed aspect of) data points. Some properties of a 
data point can be computed, measured or determined easily. These properties or characteristics are often 
referred to as \emph{features}. Beside features, there is often also some higher-level information (``quantity of interest'') 
associated with a data point. We will refer to this higher level information, or quantity of interest, as \emph{labels}. 
Many ML methods revolve around finding efficient ways to determine the label of a data point given its features. 

\begin{figure}[htbp]
	\centering
	\includegraphics[width=8cm]{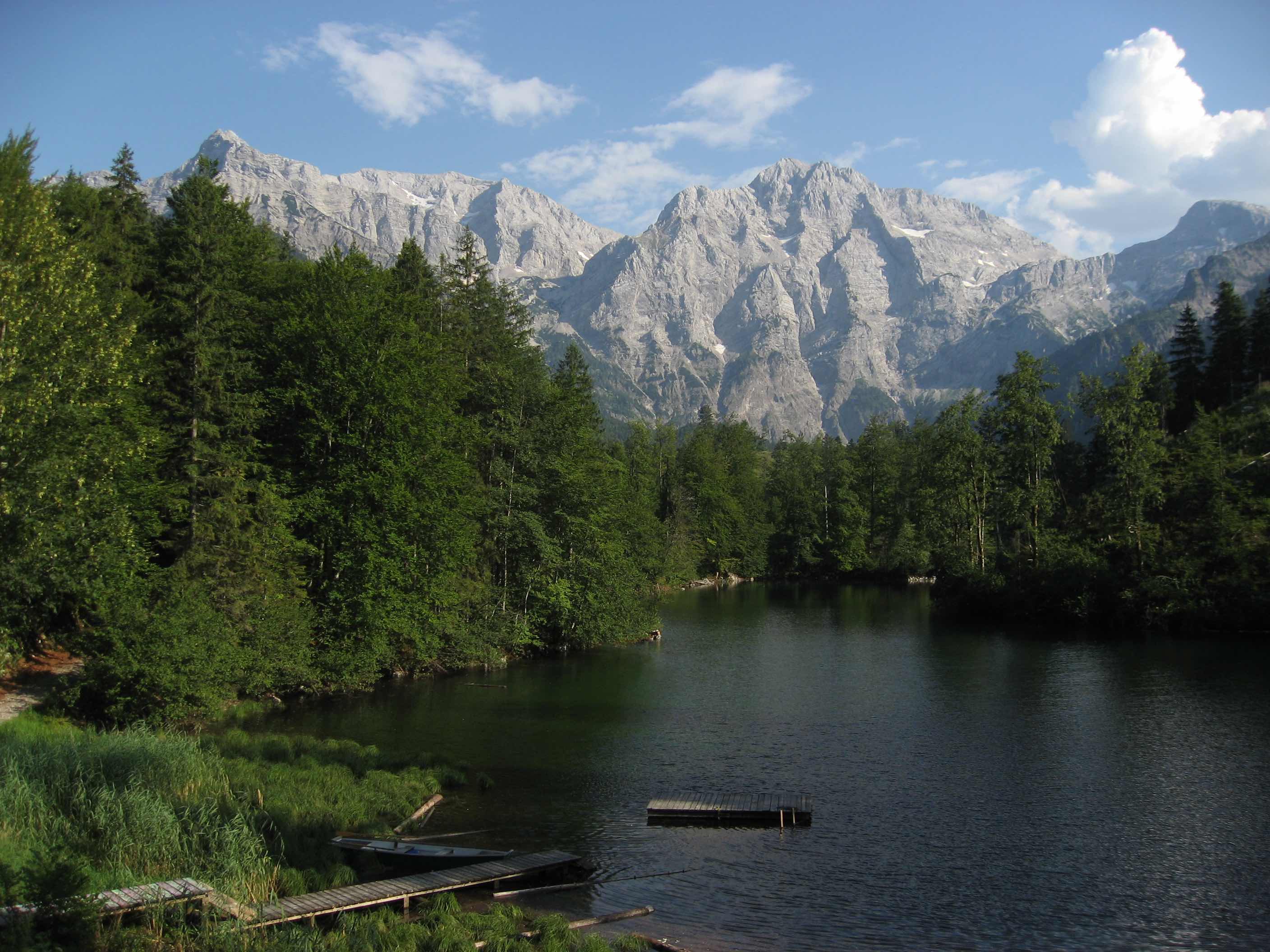}
	\caption{An image representing a data point.}
	\label{fig:image}
\end{figure}

Consider a data point represented by the snapshot depicted in Figure \ref{fig:image}. The features of this data point could be 
the red, green and blue intensities of each pixel in the image. We can stack these values into a vector $\mathbf{x} \in \mathbb{R}^{\featurelen}$ 
whose length $\featurelen$ is given by three times the number of pixels in the image. The label $y$ associated with this 
data point could be the expected hiking time to reach the mountain in the snapshot. Alternatively, we could define the 
label $y$ as the water temperature of the lake visible in the snapshot. 

The precise definition of what we use as features and labels of a data point is a design choice. 
The label is the quantity of interest for a particular application. If we are interested in developing a 
smartphone-app that predicts the hiking time given a snapshot of the mountain, we use this hiking 
time as label. However, if we are interested in developing a smartphone-app that predicts water temperature 
of a lake, we use this temperature as the label. For a given ML problem, we denote the set of all possible values that 
a label can take on by $\labelspace$. For a ML problem (method) using the choice $\labelspace = \mathbb{R}$, 
it is customary to refer to such a problem as a \emph{regression} problem (method). 

A data point is called \emph{labeled} if, besides its features $\vx$, the associated label $y$ is known.
While features are those properties or characteristics of data points that can be measured or computed easily, labels 
are difficult or costly to obtain. For the snapshot in Figure \ref{fig:image}, we can easily determine the pixel intensities as features. However, 
if the label is the water temperature of the lake depicted on the snapshot, we need to actually measure this temperature. 
Acquiring labels typically involves human labor, such as handling a water thermometer at certain locations in a lake, and 
is costly. ML methods, which have to cope with limited resources available for acquiring labels, are geared to get along 
with as little labeled data points as possible. 

Not only the label of a data point is a design choice but also what features are used to characterize a data point. In principle, 
we could use any quantity that can be easily computed or measured as a feature of a data point. Modern technology allows 
to compute a vast amount number of such quantities. 

As a case in point, consider the data point ``Alex Jung'' obtained from a person which uses a smartphone to take snapshots. 
Let us assume that Alex takes five snapshots per day on average (sometimes more, e.g., during a mountain hike). This results 
in more than $1000$ snapshots per year. Each snapshot contains around $10^{6}$ pixels. If we only use the greyscale levels of 
the pixels in all those snapshots, we would obtain more than $10^{9}$ new features per year! Modern ML applications face 
extremely high-dimensional feature vectors which calls for methods from high-dimensional statistics \cite{BuhlGeerBook,Wain2019}. 

While it might seem that ``the more features the better'', it can actually be detrimental for the performance of ML methods 
to use an excessive amount of (irrelevant) features. It is non-trivial to decide which features are most relevant for a given 
task. However, there are ML methods that allow (to some extent) to automatically learn a small number of most relevant 
features from raw data.  

\section{Hypothesis Space}

The scientific principle in Figure \ref{fig_cycle_scientific_principle} involves a hypothesis for some phenomenon 
which generates observable data. We can think of a hypothesis as a simple explanation or conception of some 
complicated phenomenon. There is a great deal of different ways to express a hypothesis. One example is a 
probability distribution which characterizes the probability of observing a particular data point. Another example 
are simple rules such as, {\rm ``if it rains in the morning, then the grass will be wet in the evening''}. Physical 
theories, such as the theory of relativity, are further examples of hypotheses. 

In general, we do not consider one single hypothesis but a whole space of alternative hypotheses. The simplest 
non-trivial hypothesis space consists of two alternative hypotheses, such as ``The earth is flat'' versus ``The earth is round''. 
We denote a hypothesis space, which consists of a set of different hypotheses, by $\mathcal{H}$. The key idea behind 
many ML methods is to choose the best hypothesis out of a large hypothesis space $\mathcal{H}$ according to some 
performance measure (see Section \ref{sec_loss_function}). 

In order to quickly search over a large hypothesis space $\mathcal{H}$, it is important to use a computer-friendly 
(representation of the) hypothesis space $\mathcal{H}$. One example of such a hypothesis space is given 
by linear predictors $h(\mathbf{x}) = \mathbf{w}^{T} \mathbf{x}$ with some weight vector $\mathbf{w} \in \mathbb{R}^{\featurelen}$. 
The resulting hypothesis space is 
\begin{equation}
\label{equ_linear_hypo_space}
\mathcal{H} \defeq \{ h^{(\vw)} (\mathbf{x}) = \mathbf{w}^{T} \mathbf{x} : \mathbf{w} \in \mathbb{R}^{\featurelen} \}. 
\end{equation} 
Each element $h^{(\mathbf{w})}$ of the hypothesis space $\mathcal{H}$ in \eqref{equ_linear_hypo_space} is a function from $\mathbb{R}^{\featurelen}$ to 
$\mathbb{R}$ which maps the feature vector $\mathbf{x}$ to the value $\mathbf{w}^{T} \mathbf{x}$. However, as indicated by the notation, 
each of the functions $h^{(\mathbf{w})}$ is fully characterized by the weight vector $\mathbf{w} \in \mathbb{R}^{\featurelen}$. Thus, we can 
parametrize the hypothesis space \eqref{equ_linear_hypo_space} using vectors $\vw$ from the Euclidean space $\mathbb{R}^{\featurelen}$. 

\begin{figure}[htbp]
 \centering
\begin{tikzpicture}[
plain/.style={
  draw=none,
  fill=none,
  },
net/.style={
  matrix of nodes,
  nodes={
    draw,
    circle,
    inner sep=10pt
    },
  nodes in empty cells,
  column sep=0.3cm,
  row sep=-9pt
  },
>=latex, 
scale=0.7
]
\matrix[net] (mat)
{
|[plain]| \parbox{1.3cm}{\centering input\\layer} & |[plain]| \parbox{1.3cm}{\centering hidden\\layer} & |[plain]| \parbox{1.3cm}{\centering output\\layer} \\
|[plain]|&  |[plain]|\\
|[plain]| & \\
& |[plain]| \\
  |[plain]| & |[plain]| \\
& & \\
  |[plain]| & |[plain]| \\
|[plain]| & |[plain]| \\
  |[plain]| & \\
|[plain]| & |[plain]| \\    };
 \draw[<-] (mat-4-1) -- node[above] {$x_{1}$} +(-2cm,0);
  \draw[<-] (mat-6-1) -- node[above] {$x_{2}$} +(-2cm,0);
    \draw[->] (mat-4-1) --  node[above]{$w_{1}$}  (mat-3-2);
      \draw[->] (mat-6-1) --  node[above]{$w_{2}$}  (mat-3-2);
        \draw[->] (mat-4-1) --  node[]{\hspace*{10mm}$w_{3}$}  (mat-6-2);
      \draw[->] (mat-6-1) --  node[above]{$w_{4}$}  (mat-6-2);
              \draw[->] (mat-4-1) --  node[]{\hspace*{10mm}$w_{5}$}  (mat-9-2);
      \draw[->] (mat-6-1) --  node[above]{$w_{6}$}  (mat-9-2);
       \draw[->] (mat-3-2) --  node[]{\hspace*{10mm}$w_{7}$}  (mat-6-3);
      \draw[->] (mat-6-2) --  node[above]{$w_{8}$}  (mat-6-3);
      \draw[->] (mat-9-2) --  node[above]{$w_{9}$}  (mat-6-3);     
      \draw[->] (mat-6-3) -- node[above] {$h^{(\vw)}(\vx)$} +(2.5cm,0);
\end{tikzpicture}
\caption{ANN representation of a predictor $h^{(\vw)}(\vx)$ which maps the 
input (feature) vector $\vx=(x_{1},x_{2})^{T}$ to a predicted label (output) $h^{(\vw)}(\vx)$.}
    \label{fig_ANN}
\end{figure}
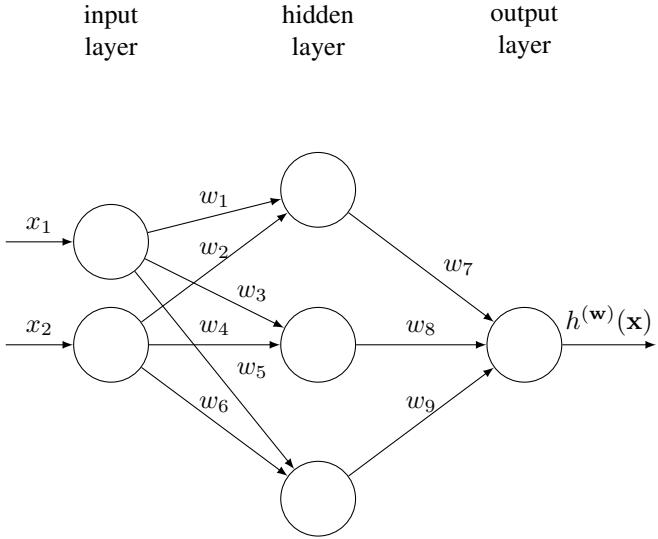

The linear space \eqref{equ_linear_hypo_space} is only one possible choice for the hypothesis space used in a ML method. 
We can also use another set of functions $h(\cdot): \featurespace \rightarrow \labelspace$ as hypothesis space. 
Decision trees define a hypothesis space using flow chart representations of the mapping $\vx \mapsto h(\vx)$ 
(see Figure \ref{fig_decision_tree}). An artificial neural network (ANN) defines a hypothesis space which consists 
of all functions that are obtained from compositions of matrix operations and simple non-linearities according to a 
network structure (see Figure \ref{fig_ANN}) . 
\begin{figure}[htbp]
\begin{minipage}{.45\columnwidth} %
\scalebox{0.8}{
\begin{tikzpicture}
  [x=2cm,->,>=stealth',level/.style={sibling distance = 4cm/#1,
   level distance = 1.5cm}] 
  \node [env] {$\| \vx-\vu \| \leq r$?}
    child { node [env] {$h(\vx) = h_{1} $}
      edge from parent node [left,align=center] {no} }
    child { node [env] {$\| \vx\!-\!\vv \|\!\leq\!r$?}
         child { node [env] {$h(\vx)\!=\!h_{2}$}
              edge from parent node [left, align=center] {no} }
         child { node [env] {$h(\vx)\!=\!h_{3}$}
              edge from parent node [right, align=center]
                {yes} }
     edge from parent node [right,align=center] {yes} };
\end{tikzpicture}
}
\end{minipage}
\begin{minipage}{.4\columnwidth} %
\hspace*{15mm}
\begin{tikzpicture}
    \tikzset{x=0.7cm} 
      \draw (-2,2) rectangle (2,-2);
      \begin{scope}
         \clip (-0.5,0) circle (1cm);
         \clip (0.5,0) circle (1cm);
         \fill[color=gray] (-2,1.5) rectangle (2,-1.5);
      \end{scope}
      \draw (-0.5,0) circle (1cm);
      \draw (0.5,0) circle (1cm);
               \draw[fill] (-0.5,0) circle [radius=0.025];
         \node [below right, red] at (-0.5,0) {$\mathcal{R}_{3}$};
                  \node [below left, blue] at (-0.7,0) {$\mathcal{R}_{2}$};
                    \node [above left] at (-0.7,1) {$\mathcal{R}_{1}$};
                  \node [left] at (-0.4,0) {$\mathbf{u}$};
                     \draw[fill] (0.5,0) circle [radius=0.025];
                     \node [right] at (0.6,0) {$\mathbf{v}$};
   \end{tikzpicture}
\end{minipage}
\caption{A decision tree represents a hypothesis $h$ which is constant on subsets $\mathcal{R}_{m}$, i.e., 
$h(\vx)\!=\!h_{m}$ for all $\vx\!\in\!\mathcal{R}_{m}$. Each subset 
$\mathcal{R}_{m}\!\subseteq\!\featurespace$ corresponds to a leaf node in the decision tree.}
\label{fig_decision_tree}
\end{figure}
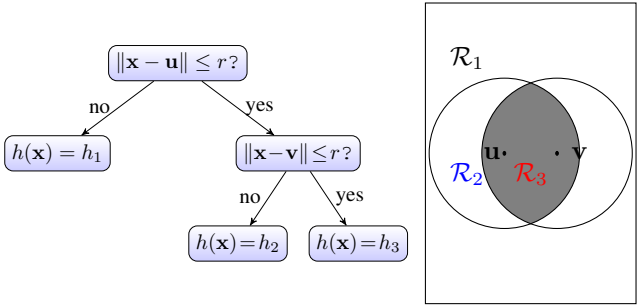 

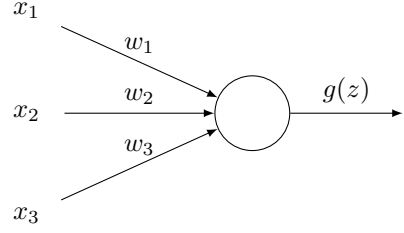
\begin{figure}[htbp]
\centering
\begin{tikzpicture}[
plain/.style={
  draw=none,
  fill=none,
  },
net/.style={
  matrix of nodes,
  nodes={
    draw,
    circle,
    inner sep=10pt
    },
  nodes in empty cells,
  column sep=2cm,
  row sep=-9pt
  },
>=latex
]
\matrix[net] (mat)
{
|[plain]| &|[plain]| \\
|[plain]|& |[plain]| \\
  |[plain]| & |[plain]| \\
|[plain]|& & |[plain]|\\
|[plain]|& |[plain]| \\
|[plain]| & |[plain]| \\
  |[plain]| &|[plain]| \\  };
   \draw[->] (mat-2-1) node[]{$x_{1}$} --  node[above]{$w_{1}$}  (mat-4-2);
        \draw[->] (mat-4-1) node[]{$x_{2}$}  --  node[above]{$w_{2}$}  (mat-4-2) ;
      \draw[->] (mat-6-1) node[]{$x_{3}$}--  node[above]{$w_{3}$}  (mat-4-2);
      \draw[->] (mat-4-2) -- node[above] {$g(z)$} +(2cm,0);
\end{tikzpicture}
\caption{Each single neuron of the ANN depicted in Figure \ref{fig_ANN} 
implements a weighted summation $z=\sum_{i} w_{i} x_{i}$ of its inputs 
$x_{i}$ followed by applying a non-linear activation function $g(z)$.}
\label{fig_activate_neuron}
\end{figure}

The choice for the hypothesis space $\hypospace$ has to balance two conflicting requirements: 
\begin{itemize} 
\item It has to be sufficiently large (or rich) such that it contains a predictor map $\hat{h} \in \hypospace$ 
that is able to represent (approximate) the underlying relation between the features and the label of a data point.  
\item It has to be sufficiently small (compact) such that it can be efficiently searched over to 
find good predictors during a training phase. This requirement typically necessitates that an arbitrary 
maps $h(\vx)$ contained in $\hypospace$ can be evaluated (computed) efficiently \cite{Austin2018}. 
\end{itemize}
	
\section{Loss Function}
\label{sec_loss_function}
To find good predictor maps we need some quality measure that 
allows assess a given predictor function $h \in \mathcal{H}$. Many ML methods 
use the concept of a loss function $\mathcal{L}((\vx,y),h)$ that represents the loss 
(error) incurred by using the predictor $h$ to predict the label $y$ of a data point 
with features $\mathbf{x}$. 

Just like feature space, label space and hypothesis space, also the loss function is 
a design parameter. In principle, we can use any function $\mathcal{L}: \featurespace \times \labelspace \times \hypospace \rightarrow \mathbb{R}$ 
that maps a data point $(\vx,y)$ and hypothesis $h\in \hypospace$ a number 
 $\mathcal{L}((\vx,y),h)$ that represents the loss of using the predictor map $h$ to predict 
 the label $y \in \labelspace$ of a data point with features $\vx \in \featurespace$. 

Popular choices are 
\begin{itemize} 
\item the squared error loss 
\begin{equation} 
\label{equ_def_squared_error_loss}
\mathcal{L}= (y - \underbrace{h(\vx)}_{\hat{y}})^{2}, 
\end{equation} 
for regression problems with label space $\labelspace = \mathbb{R}$.  
\item the logistic loss 
\begin{equation} 
\label{equ_def_logistic_loss}
\mathcal{L}= - \log (1+\exp(-y h(\vx) )), 
\end{equation} 
for binary classification problems with label space $\labelspace= \{-1,1\}$. 
\item the Huber loss 
\begin{equation} 
\label{equ_def_huber_loss}
\mathcal{L} = \begin{cases} (1/2) (y-h(\vx))^{2} & \mbox{ for } |y-h(\vx)| \leq  c \\ 
c (|y-h(\vx)| - c/2) & \mbox{ else. }\end{cases}
\end{equation} 
with some tuning parameter $c$ controlling the threshold of whether the error for a given data point should follow the squared loss or the absolute loss which is more appropriate for outliers (note that if $c$ is selected as a large value, the Huber loss would be equivalent to squared loss divided by two). The Huber loss can be used for label space $\labelspace = \mathbb{R}$.   
\end{itemize}

The choice of loss functions is guided by statistical and computational aspects. 
Learning a predictor by minimizing the squared error loss \eqref{equ_def_squared_error_loss} 
amounts to \emph{maximum likelihood estimation} if the labels are modeled as 
\begin{equation} 
\label{equ_AWGN_model}
y = \bar{h}(\vx)+\varepsilon.
\end{equation} 
The model \eqref{equ_AWGN_model} involves some true predictor $\bar{h}$ (which is unknown) 
and a random variable $\varepsilon \sim \mathcal{N}(0,1)$ which covers any modeling and 
measurement (labeling) errors. Thus, if the model \eqref{equ_AWGN_model} accurately 
describes the observed labels $y$ of data points (which can be considered as statistically independent), 
the squared error loss \eqref{equ_def_squared_error_loss} is a statistically optimal choice. 

Using the logistic loss \eqref{equ_def_logistic_loss} amounts to maximum likelihood 
estimation when the labels $y \in \{-1,1\}$ are modelled as random variables with probability 
\begin{equation}
\label{equ_logistic_model}
{\rm Prob} \{y =1 \} = 1/(1+\exp(-y \bar{h}(\vx)))
\end{equation} 
with some true predictor $\bar{h}$ (which is unknown). 

Aside from their statistical properties, loss functions differ in their computational properties. 
The squared error loss \eqref{equ_def_squared_error_loss} and the logistic loss \eqref{equ_def_logistic_loss} 
are computationally attractive since they amount to minimizing a differentiable and convex function. 
Such smooth convex optimization problems can be solved efficiently via (stochastic) gradient descent methods \cite{nestrov04,HazanOCO}. 

Sometimes it is beneficial to use non-smooth (non-differentiable) loss functions. In applications where 
few data points are severely corrupted (e.g., by a broken device)  it is beneficial to use the Huber loss 
\eqref{equ_def_huber_loss} \cite{HuberRobustBook}. Optimizing non-smooth functions is typically more 
challenging, requiring more computational resources, compared to optimizing smooth functions. 

\section{Putting Together The Pieces}

Many ML method are obtained by combining particular choices for feature space $\featurespace$ 
and label space, hypothesis space $\mathcal{H}$ and loss function $\mathcal{L}$. One of the most 
basic and widely used ML methods is \emph{linear regression}. 

Linear regression chooses an optimal linear predictor out of the hypothesis space \eqref{equ_linear_hypo_space} 
by minimizing the average squared error loss, or mean squared error, 
\begin{align} 
\label{equ_mse}
& (1/\samplesize) \sum_{\sampleidx=1}^{\samplesize} \big( y^{(\sampleidx)} - h\big(\mathbf{x}^{(\sampleidx)}\big) \big)^{2}   \nonumber  \\
& = (1/\samplesize) \sum_{\sampleidx=1}^{\samplesize} \big( y^{(\sampleidx)} - \mathbf{w}^{T} \mathbf{x}^{(\sampleidx)} \big)^{2}. 
\end{align} 
The average squared error loss is obtained by comparing the prediction $h(\vx^{(\sampleidx)}\big)$ of 
the linear predictor $h(\mathbf{x}) = \mathbf{w}^{T} \vx$ to the true label $y^{(\sampleidx)}$ of a data point 
with features $\mathbf{x}^{(\sampleidx)}$. Note that the criterion \eqref{equ_mse} requires $\samplesize$ 
labeled data points with features $\vx^{(\sampleidx)}$ and known labels $y^{(\sampleidx)}$. 

\begin{figure}[htbp]
\begin{center}
\begin{tikzpicture}
 \tikzset{x=3cm,y=0.1cm,every path/.style={>=latex},node style/.style={circle,draw}}
    \csvreader[ head to column names,%
                late after head=\xdef\aold{\x}\xdef\eold{\y}\xdef\hatold{\yhat},,
                after line=\xdef\aold{\x}\xdef\eold{\y}\xdef\hatold{\yhat}]%
                {cleandata.csv}{}
                {
                \draw [line width=0.4mm] (\aold, \hatold)-- (\x,\yhat) ;
    }
  \csvreader[ head to column names,%
                late after head=\xdef\aold{\x}\xdef\eold{\y}\xdef\hatold{\yhat},,
                after line=\xdef\aold{\x}\xdef\eold{\y}\xdef\hatold{\yhat}]%
                {cleandata.csv}{}
                {\draw [line width=0mm] (\aold, \eold)  (\x,\y) node  {$\circ$};
    }

          \draw[->] (-1.0,0) -- (1.0,0);
      \node [anchor=south] at (1,0) {feature $x$};
      \draw[->] (0,-20) -- (0,20);   
            \node [anchor=south] at (0,20.4) {label $y$};
   \end{tikzpicture}
\end{center}
  \caption{A data set consisting of labeled data points $(x^{(i)},y^{(i)}$ (depicted as ``$\circ$'') and 
  the linear predictor $h(x)=w x$ (solid line) obtained by minimizing the average squared error \eqref{equ_mse}.}
  \label{fig_scalarlinreg}
  \vspace*{-3mm}
\end{figure}
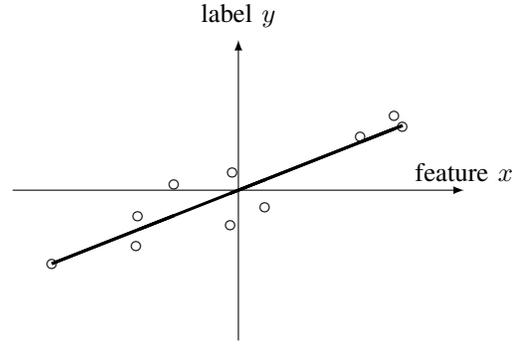

In Figure \ref{fig_scalarlinreg}, we depict a set of labeled data points which are used to 
learn a linear predictor by minimizing the average squared error \eqref{equ_mse}. As 
hinted at in Section \ref{sec_loss_function}, learning a predictor by minimizing the average 
squared error \eqref{equ_mse} is statistically optimal if the labels and features are related 
by the additive Gaussian noise model \eqref{equ_AWGN_model}. 

%
 
For some datasets the model \eqref{equ_mse} does not accurately reflect the relation between 
features and labels. In particular, some data sets contain outliers which have fundamentally 
different properties compared to the bulk of (clean) data points. We can think of outliers as 
being the result of exceptional events such as failure of hardware (e.g., broken sensing device).  

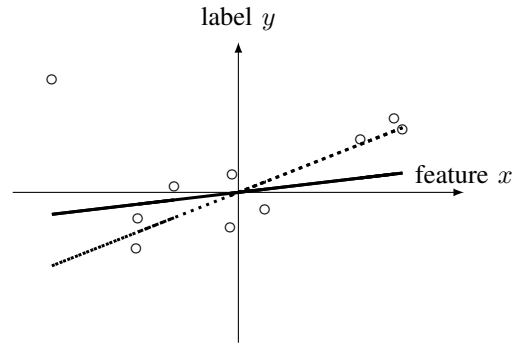
\begin{figure}[htbp]
\begin{center}
\begin{tikzpicture}
 \tikzset{x=3cm,y=0.1cm,every path/.style={>=latex},node style/.style={circle,draw}}
    \csvreader[ head to column names,%
                late after head=\xdef\aold{\x}\xdef\eold{\y}\xdef\hatold{\yhat},,
                after line=\xdef\aold{\x}\xdef\eold{\y}\xdef\hatold{\yhat}]%
                {corrupteddata.csv}{}
                {\draw [line width=0mm] (\aold, \eold)  (\x,\y) node  {$\circ$};
                \draw [line width=0.4mm] (\aold, \hatold)-- (\x,\yhat) ;
    }
    
        \csvreader[ head to column names,%
                late after head=\xdef\aold{\x}\xdef\eold{\y}\xdef\hatold{\yhat},,
                after line=\xdef\aold{\x}\xdef\eold{\y}\xdef\hatold{\yhat}]%
                {cleandata.csv}{}
                {
                \draw [line width=0.4mm, dotted] (\aold, \hatold)-- (\x,\yhat) ;
    }
          \draw[->] (-1.0,0) -- (1.0,0);
      \node [anchor=south] at (1,0) {feature $x$};
      \draw[->] (0,-20) -- (0,20);   
            \node [anchor=south] at (0,20.4) {label $y$};
\end{tikzpicture}
\end{center}
  \caption{Corrupted data set (depicted as ``$\circ$'') which is the same as in Figure \ref{fig_scalarlinreg} except for the left-most data point. 
  The solid line represents the linear predictor $h(x)=w x$ (solid line) obtained by minimizing the average squared error \eqref{equ_mse} on 
  the corrupted data set. The dotted line indicated the predictor obtained from the clean data set (solid line in \eqref{fig_scalarlinreg}).}
  \label{fig_scalarlinreg_corr}
  \vspace*{-3mm}
\end{figure}

It turns out that learning a predictor by minimizing the squared error loss \eqref{equ_mse} is not 
robust against outliers. We illustrate this non-robustness in Figure \ref{fig_scalarlinreg_corr} 
which depicts a data set that is obtained by corrupting one single data point form the data set 
shown in Figure \ref{fig_scalarlinreg}. Minimizing the average squared error loss on the perturbed 
data set results in a different linear predictor (solid line in Figure \ref{fig_scalarlinreg_corr}) than 
for the clean data set (dotted line in Figure \ref{fig_scalarlinreg_corr}). Thus, if only one single 
data point is corrupted, minimizing the squared error loss results in significantly different predictors.

\begin{figure}[htbp]
\begin{center}
\begin{tikzpicture}
 \tikzset{x=3cm,y=0.1cm,every path/.style={>=latex},node style/.style={circle,draw}}
    \csvreader[ head to column names,%
                late after head=\xdef\aold{\x}\xdef\eold{\y}\xdef\hatold{\yhat},,
                after line=\xdef\aold{\x}\xdef\eold{\y}\xdef\hatold{\yhat}]%
                {corrupteddata_huber.csv}{}
                {\draw [line width=0mm] (\aold, \eold)  (\x,\y) node  {$\circ$};
                \draw [line width=0.4mm] (\aold, \hatold)-- (\x,\yhat) ;
    }
    
        \csvreader[ head to column names,%
                late after head=\xdef\aold{\x}\xdef\eold{\y}\xdef\hatold{\yhat},,
                after line=\xdef\aold{\x}\xdef\eold{\y}\xdef\hatold{\yhat}]%
                {cleandata_huber.csv}{}
                {
                \draw [line width=0.4mm, dotted] (\aold, \hatold)-- (\x,\yhat) ;
    }
          \draw[->] (-1.0,0) -- (1.0,0);
      \node [anchor=south] at (1,0) {feature $x$};
      \draw[->] (0,-20) -- (0,20);   
            \node [anchor=south] at (0,20.4) {label $y$};
\end{tikzpicture}
\end{center}
  \caption{Corrupted data set (depicted as ``$\circ$'') which is the same as in Figure \ref{fig_scalarlinreg} except for the left-most data point. 
  The solid line represents the linear predictor $h(x)=w x$ (solid line) obtained by minimizing the average Huber loss \eqref{equ_def_huber_loss} on 
  the corrupted data set. The dotted line indicated the predictor obtained from minimizing the average Huber loss on the clean data set (depicted by the 
  circles in Figure \ref{fig_scalarlinreg}).}
  \label{fig_scalarlinreg_corr_Huber}
  \vspace*{-3mm}
\end{figure}
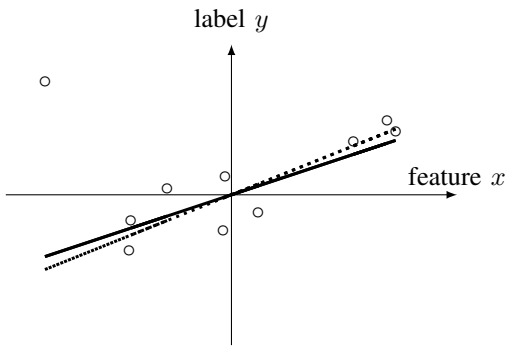

In order to obtain more robustness against few outliers in the data set we might use 
the Huber loss \eqref{equ_def_huber_loss}. Figure \ref{fig_scalarlinreg_corr_Huber} 
depicts the same corrupted data set as used in Figure \ref{fig_scalarlinreg_corr}. The 
solid line depicts the linear predictor obtained by minimizing 
the average Huber loss incurred on the corrupted data set, while the dotted 
line indicated the linear predictor obtained by minimizing the average Huber loss on the 
clear data set (depicted as circles in Figure \ref{fig_scalarlinreg}). 

By comparing Figure \ref{fig_scalarlinreg_corr_Huber} with Figure \ref{fig_scalarlinreg_corr}, 
we conclude that using the Huber loss \eqref{equ_def_huber_loss} instead of the squared 
error loss \eqref{equ_def_squared_error_loss} results in a more robust ML method. 
However, this comes at the price of a more challenging optimization problem since 
the Huber loss is non-differentiable. 

\section*{Acknowledgments}
We thank the students of the Aalto courses ``Machine Learning: Basic Principles'', ``Artificial Intelligence'' 
and ``Machine Learning with Python'' for their constructive and critical feedback. This feedback was instrumental 
for the author to learn how to teach ML. Teaching assistant Shaghayegh Safar helped with a careful review 
of an early version of the draft. 
\bibliographystyle{IEEEtran}
\bibliography{/Users/junga1/Literature.bib}

\end{document}

%% file: mlbpbook_macros.tex
\newcommand{\vx}[0]{{\bf x}}
\newcommand{\vv}[0]{{\bf v}}
\newcommand{\vu}[0]{{\bf u}}

\newcommand{\vw}[0]{{\bf w}}

\newcommand{\bmx}[0]{\begin{bmatrix}}
\newcommand{\emx}[0]{\end{bmatrix}}

\newcommand\defeq{:=}

\newcommand{\featuredim}{n}
\newcommand{\samplesize}{m}
\newcommand{\sampleidx}{i}

\newcommand{\hypospace}{\mathcal{H}}

\newcommand{\featurespace}{\mathcal{X}}
\newcommand{\labelspace}{\mathcal{Y}}